\documentclass{article}

\usepackage{microtype}
\usepackage{graphicx}
\usepackage{subfigure}
\usepackage{booktabs} 

\usepackage{hyperref}

\usepackage[accepted]{icml2021mod}

\usepackage{amsmath}
\usepackage{algorithm}
\usepackage{algorithmic}
\usepackage{amssymb}
\usepackage{amsthm}
\usepackage{graphicx}
\usepackage{subfigure}
\usepackage{adjustbox}
\usepackage{hhline}
\usepackage{makecell}
\usepackage{enumitem}
\usepackage{soul}           
\usepackage{array, multicol, multirow}
\usepackage{booktabs}
\usepackage{hhline}
\usepackage[dvipsnames]{xcolor}

\newcommand{\F}{\mathbb{F}}
\newcommand{\Q}{\mathbb{Q}}
\newcommand{\hF}{\hat{\mathbb{F}}}
\newcommand{\hQ}{\hat{\mathbb{Q}}}

\newcommand{\cx}{\mathcal{X}}
\newcommand{\cy}{\mathcal{Y}}

\newcommand{\bX}{\boldsymbol{X}}

\newcommand{\bY}{\boldsymbol{Y}}

\newcommand{\FF}{\mathbb{F}}

\icmltitlerunning{Uncertainty Toolbox}

\begin{document}
\twocolumn[
\icmltitle{Uncertainty Toolbox: an Open-Source Library for Assessing, Visualizing, \\and Improving Uncertainty Quantification}



\icmlsetsymbol{equal}{*}

\begin{icmlauthorlist}
\icmlauthor{Youngseog Chung}{ri,mld}
\icmlauthor{Ian Char}{mld}
\icmlauthor{Han Guo}{lti}
\icmlauthor{Jeff Schneider}{ri}
\icmlauthor{Willie Neiswanger}{stf}
\end{icmlauthorlist}

\icmlaffiliation{ri}{Robotics Institute,
}
\icmlaffiliation{mld}{Machine Learning Department,
}

\icmlaffiliation{lti}{Language Technologies Institute, Carnegie Mellon University, Pittsburgh, Pennsylvania, USA}
\icmlaffiliation{stf}{Computer Science Department, Stanford University, California, USA}

\icmlcorrespondingauthor{Youngseog Chung}{youngsec@cs.cmu.edu}

\icmlkeywords{Machine Learning, ICML, uncertainty quantification, calibration, toolbox}

\vskip 0.3in
]



\printAffiliationsAndNotice{}  

\vspace{-3mm}
\section{Introduction} 
\label{sec:introduction}
\vspace{-1mm}

%
As machine learning (ML) systems are increasingly deployed on an array of
high-stakes tasks, there is a growing need to robustly quantify their predictive uncertainties.
Uncertainty quantification (UQ) in
machine learning generally refers to the task of quantifying the confidence of a given prediction,
and this measure of confidence can be especially crucial in a variety of downstream applications, including
Bayesian optimization
\citep{jones1998efficient, shahriari2015taking},
model-based reinforcement learning \citep{malik2019calibrated, yu2020mopo},
and in
high-stakes prediction settings where errors 
incur large costs \citep{wexler2017computer, rudin2019stop}.

UQ is often performed via \emph{distributional predictions} (in contrast with \emph{point predictions}).
Hence, 
given inputs $x \in \mathcal{X}$  and targets $y \in \mathcal{Y}$, 
one common goal in UQ is to 
approximate the true conditional distribution of $y$ given $x$.
In the supervised setting where we only have access to a
limited data sample, 
we are then faced with the question,
``how can one verify whether a distributional prediction is close to the true distribution using only a finite dataset?''
Many works in UQ tend to be disjoint in the evaluation metric utilized,
which sends divided signals about which metrics \textit{should}
or \textit{should~not} be used.
For example, some works report likelihood on a test set
\citep{lakshminarayanan2017simple, detlefsen2019reliable, zhao2020individual}, some works use other proper scoring rules \citep{maciejowska2016probabilistic, askanazi2018comparison, bowman2020uncertainty, bracher2021evaluating}, while others focus on calibration metrics \citep{kuleshov2018accurate, cui2020calibrated}.
Further, with disparate implementations for each metric, 
it is often the case that reported numerical results are not directly 
comparable across different works, even if a similar metric is used.

To address this, we present \emph{Uncertainty Toolbox}:
an open-source python library that helps to assess, visualize, and improve UQ. 
There are other libraries such as 
Uncertainty Baselines~\citep{nado2021uncertainty} and Robustness Metrics~\citep{djolonga2020robustness}
that focus on aspects of UQ in the \textit{classification} setting. 
Uncertainty Toolbox focuses on the \textit{regression} setting and 
additionally aims to provide user-friendly utilities 
such as visualizations, a glossary of terms,
and an organized collection of key paper references.

We begin our discussion by first introducing the
contents of Uncertainty Toolbox.
We then provide an overview of evaluation metrics in UQ.
Afterwards, we demonstrate the functionalities of the toolbox 
with a case study where we train probabilistic neural networks (PNNs)
\citep{nix1994estimating, lakshminarayanan2017simple}
with a set of different loss functions, and evaluate the 
resulting trained models using metrics and visualizations in the toolbox.
This case study shows that certain evaluation metrics
shed light on different aspects of UQ performance, and makes the case for
using a suite of metrics for a comprehensive evaluation.

\vspace{-3mm}
\section{Toolbox Contents}
\vspace{-1mm}

Uncertainty Toolbox comprises four main functionalities, which we detail below.
\vspace{-4mm}
\paragraph{Evaluation Metrics}
The toolbox provides implementations for a suite of evaluation metrics.
The main categories of metrics are: 
calibration, group calibration, sharpness, and proper scoring rules.
We discuss each of these metric types in the following
section (Section~\ref{sec:metrics}).

\vspace{-4mm}
\paragraph{Recalibration}
We further implement recalibration methods
that leverage isotonic regression \citep{kuleshov2018accurate}.
Concretely, recalibration aims to improve the average calibration
(defined in Eq.~(\ref{eq:avg_cali})) of distributional predictions. 

\vspace{-4mm}
\paragraph{Visualizations}
The toolbox offers a range of easy-to-use visualization utilies to help in inspecting and evaluating UQ quality.
These plotting utilities focus on visualizing the predicted distribution,
calibration, and prediction accuracy.

\vspace{-4mm}
\paragraph{Pedagogy}
For those unfamiliar with the area of predictive UQ, we provide a glossary that communicates the core concepts in this area, and maintain a paper list which organizes some of the key papers in the field.

We hope the toolbox serves as an intuitive guide
for those unfamiliar but interested in utilizing UQ, 
and as a practical tool and point of reference for those active in UQ research.

\textbf{Uncertainty Toolbox is available at the following page:}
{\url{https://github.com/uncertainty-toolbox/uncertainty-toolbox}}.



\vspace{-3mm}
\section{Evaluation Metrics in Predictive UQ}
\label{sec:metrics}
\vspace{-1mm}
To summarize the notation and setting:
\textbf{X, Y} denote random variables; $x, y$ denote realized values; and $\mathcal{X, Y}$ denote sets of possible values. 
Further, 
for any random variable, 
we denote the true CDF as
${\FF}$,
its inverse (i.e. the quantile function) as
$\Q$,
the corresponding density function as
${f}$,
and the space of distributions as $\mathcal{F}$.
Estimates of these true functions will be denoted with a hat, e.g. $\hat{\mathbb{F}}$ and $\hat{f}$.
Lastly, we consider the regression setting where $\cy \subset \mathbb{R}$ and $\mathcal{X} \subset \mathbb{R}^n$.

Many recent works have focused on evaluation metrics involving
notions of \textit{calibration} and \textit{sharpness}
\citep{gneiting2007probabilistic, guo2017calibration, kuleshov2018accurate, song2019distribution, tran2020methods,zhao2020individual,  fasiolo2020fast, cui2020calibrated}.
Calibration in the regression setting is defined in terms of quantiles, and broadly speaking, it requires that 
the probability of observing the target random variable below a predicted $p$\textsuperscript{th} quantile 
is equal to the \textit{expected probability} $p$, for all $p \in (0, 1)$.
We refer to the former quantity as the \textit{observed probability} (also referred to as empirical probability) and denote it 
$p^{\text{obs}}(p)$, for an expected probability $p$.
Calibration requires $p^{\text{obs}}(p) = p$, $\forall p \in (0,1)$.
From this generic statement, we can describe different notions of calibration
based on how $p^{\text{obs}}$ is defined.

The most common form of calibration is \textbf{average calibration}, where $\hQ_p(x)$ is the estimated $p$\textsuperscript{th} quantile of $\textbf{Y} | x$,
\vspace{-1mm}
\begin{align} \label{eq:avg_cali}
    p^{\text{obs}}_{avg}(p) := \mathbb{E}_{x \sim \F_{\textbf{X}}}[\mathbb{F}_{\textbf{Y}|x}(\hQ_p(x))], \hspace{2mm} \forall p\in (0, 1),
\end{align} 
i.e. the probability of observing the target below the quantile prediction, 
\textit{averaged over $\F_{\textbf{X}}$}, is equal to $p$. 
Average calibration is often referred to simply as ``calibration''
\citep{kuleshov2018accurate, cui2020calibrated},
and it is amenable to estimation in finite datasets, as follows.
Given a dataset $D = \{(x_i, y_i)\}_{i=1}^{N}$, we can estimate $p^{\text{obs}}_{avg}(p)$ with
$\hat{p}^{\text{obs}}_{avg}(D, p) =
\frac{1}{N}\sum_{i=1}^{N} \mathbb{I}\{y_i \leq \hQ_p(x_i)\}$.
%
%
%
%
The degree of error in average calibration is commonly measured by \textit{expected calibration error} \cite{guo2017calibration, tran2020methods, cui2020calibrated},
$\text{ECE}(D, \hQ) = \frac{1}{m}\sum_{j=1}^{m} \left | \hat{p}^{\text{obs}}_{avg}\left(D, p_j\right) - p_j \right |$,
where $p_j$ is a range of expected probabilities of interest.
Note that if our quantile estimate achieves average calibration then $\hat{p}^{\text{obs}}_{avg} \rightarrow p$ (and thus ECE $\rightarrow 0$) as $N\rightarrow\infty$, $\forall p \in (0, 1)$.

It may be possible to have an uninformative, yet average calibrated model. For example,
quantile predictions that match the true \textit{marginal} quantiles of $\F_{\textbf{Y}}$ 
will be average calibrated, but will hardly be useful since they do not depend on the input $x$.
Therefore, the notion of \textbf{sharpness} is also considered, which quantifies the concentration of distributional predictions \citep{gneiting2007probabilistic}.
For example, in predictions that parameterize a Gaussian, 
the variance of the predicted distribution is often taken as a measure of sharpness.
There generally exists a tradeoff between average calibration and sharpness \citep{murphy1973new, gneiting2007probabilistic}.

Recent works have suggested a notion of calibration stronger than average calibration, called adversarial group calibration \citep{zhao2020individual}.
This stems from the notion of \textbf{group calibration} \citep{kleinberg2016inherent, hebert2017calibration}, which prescribes measurable subsets 
$\mathcal{S}_i \subset \cx$ s.t. $P_{x \sim \F_{\textbf{X}}}(x \in \mathcal{S}_i) > 0$, $i=1,\dots,k$, 
and requires the predictions to be average calibrated within each subset.
Adversarial group calibration then
requires average calibration for \textit{any subset of $\cx$ with non-zero measure}.
Denote $\bX_{\mathcal{S}}$ as a random variable that is conditioned on being in the set $\mathcal{S}$.
For \textbf{adversarial group calibration}, the observed probability is
\vspace{-1mm}
\begin{align} \label{eq:adversarial-group-calibration}
\begin{split}
    %
    %
    p^{\text{obs}}_{adv}(p) &:= \hspace{2mm} \mathbb{E}_{x \sim \F_{\textbf{X}_{\mathcal{S}}}}[\mathbb{F}_{\textbf{Y}|x}(\hQ_p(x))],\\
    \hspace{2mm} \forall p \in (0, 1),
    \hspace{2mm} &\forall \mathcal{S} \subset \cx \text{  s.t.  } P_{x \sim \F_{\textbf{X}}} (x \in \mathcal{S}) > 0.
\end{split}
\end{align}
With a finite dataset, we can measure a proxy of adversarial group calibration by measuring average calibration within all subsets of the data with sufficiently many points.

An alternative but widely used family of evaluation metrics is \textbf{proper scoring rules} \citep{gneiting2007strictly}. 
Proper scoring rules are summary statistics of overall performance of a
distributional prediction, and are defined such that the true underlying distribution optimizes the expectation of the scoring rule.
Given a scoring rule $S(\hF, (x,y))$, where $x\sim\FF_{\bX}$, $y \sim\FF_{\bY|x}$, the expectation of the scoring rule is
$S(\hF, \FF) = \mathbb{E}_{\bX, \bY}[S(\hF, (x,y))]$, 
and $S$ is said to be a proper scoring rule if
$S(\FF, \FF) \geq S(\hF, \FF), \forall \hF \in \mathcal{F}$.

There are a variety of proper scoring rules, based on the
representation of the distributional prediction. 
Since these rules consider both calibration and sharpness together 
in a single value~\citep{gneiting2007probabilistic}, 
they also serve as optimization objectives for UQ. 
For example, the logarithmic score 
is a popular proper scoring rule for density predictions \citep{lakshminarayanan2017simple, pearce2018uncertainty, detlefsen2019reliable}, and it is used as a loss function
via \textit{negative log-likelihood} (\textbf{NLL}).
The \textbf{check score} is widely used for quantile predictions and also known as the \textit{pinball loss}.
The \textbf{interval score} is commonly used for prediction intervals (a pair of quantiles with a prescribed expected coverage), 
and the continuous ranked probability score (\textbf{CRPS}) 
is popular for CDF predictions
\footnote{Proper scoring rules are usually \textit{positively oriented} (i.e. greater value is more desirable), 
and their negative is taken as a loss function to minimize.
In our work, we always report proper scoring rules in their \textit{negative orientation} (i.e. as a loss).}.
We refer the reader to \citet{gneiting2007strictly} for the 
definition of each scoring rule.




Given the wide range of metrics available, one might naturally ask,
``is there one metric to rule them all?''
Previous work has investigated some aspects of this question.
For example, \citet{chung2020beyond} noted the mismatch between
the check score and average calibration, and
\citet{gneiting2007strictly} and \citet{bracher2021evaluating} point out
cases in which disagreements can occur between some scoring rules.
Still, whether there exists a golden metric in UQ is an open research problem.
We instead suggest that there is virtue in inspecting various metrics
simultaneously, which  is made easy by Uncertainty Toolbox, as we show below.

\vspace{-3mm}
\section{Case Study on Training, Evaluating PNNs}
\vspace{-1mm}
\label{sec:case_study}
To demonstrate the capabilities of Uncertainty Toolbox,
we provide a case study on training PNNs with various loss objectives,
and use the toolbox to examine the results. 

A PNN is a neural network that assumes a conditional Gaussian for the
predictive distribution, thus for any input point $x$, outputs
an estimate of the mean and the covariance, $\hat{\mu}(x)$, $\hat{\Sigma}(x)$.
This NN structure has been proposed as early as \citet{nix1994estimating},
but it has been popularized as a UQ method in deep learning by
\citet{lakshminarayanan2017simple},
and it remains one of the most popular UQ methods to date.

The standard method of training PNNs is to optimize the logarithmic score, i.e. NLL loss. 
However, based on the training principle, ``optimize a proper score to improve UQ quality''
\citep{lakshminarayanan2017simple} (also referred to as ``optimum score estimation'' by \citet{gneiting2007strictly}), 
we can in fact optimize many more proper scoring rules. 
In this study, we train PNNs using several different methods by optimizing with respect to either NLL, CRPS, check score, or interval score.
Afterwards, we assess the predictive UQ quality with Uncertainty Toolbox.
We summarize main details of the experiment below (full details in Appendix~\ref{app:exp_details}).

\vspace{-4mm}
\paragraph{Dataset} The data was generated with a mean function $y = \sin(x/2) + x\cos(0.8x)$ and heteroscedastic Gaussian noise was added to  
generate the observations, $y$, for each input $x \sim \text{unif} [-10, 10]$.
The train, validation and test splits 
consisted of $200, 100, 100$ points, respectively.

\vspace{-4mm}
\paragraph{Training} A separate model was trained for each loss function with full batch gradient decent and learning rate $1e^{-3}$, for 2000 epochs while tracking the validation loss.
To optimize the check and interval scores, 
a batch of 30 expected probabilities $p_i \sim \text{unif}(0, 1)$ was selected 
and the scores for each $p_i$ were summed to compute the loss \citep{tagasovska2019single, chung2020beyond}.
All reported results are based on the model with best validation loss.

\begin{figure}[t!]
    \centering
    \includegraphics[width=0.45\textwidth]{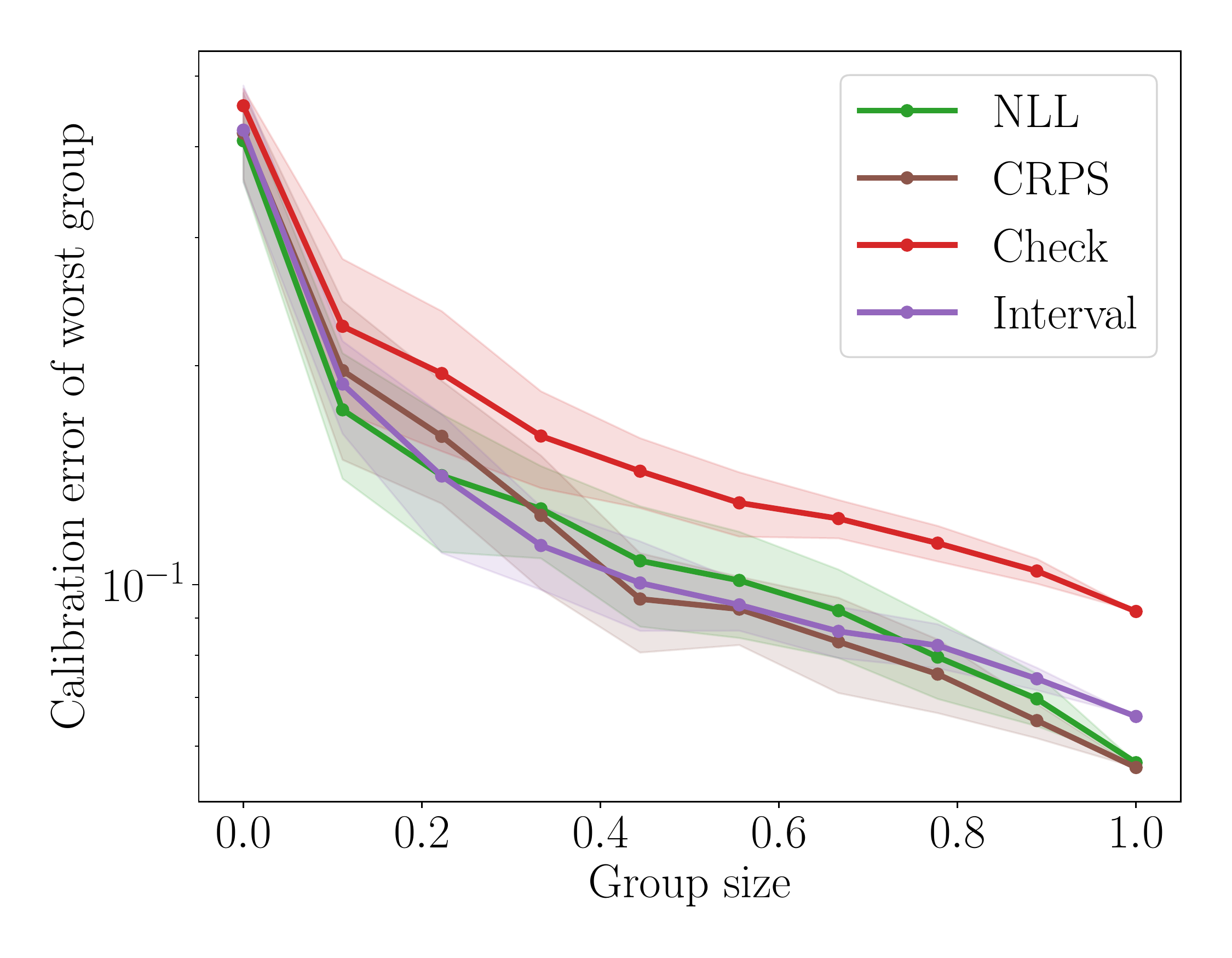}
    \vspace{-5mm}
    \caption{
    \textbf{Adversarial Group Calibration.}
    Group size refers to proportion of test dataset size,
    and the shades represent $\pm1$ standard error 
    for the calibration error of the worst group.
    \label{fig:agc}}
    \vspace{-3mm}
\end{figure}

\begin{figure*}[ht!]
    \centering
    \includegraphics[width=\textwidth]{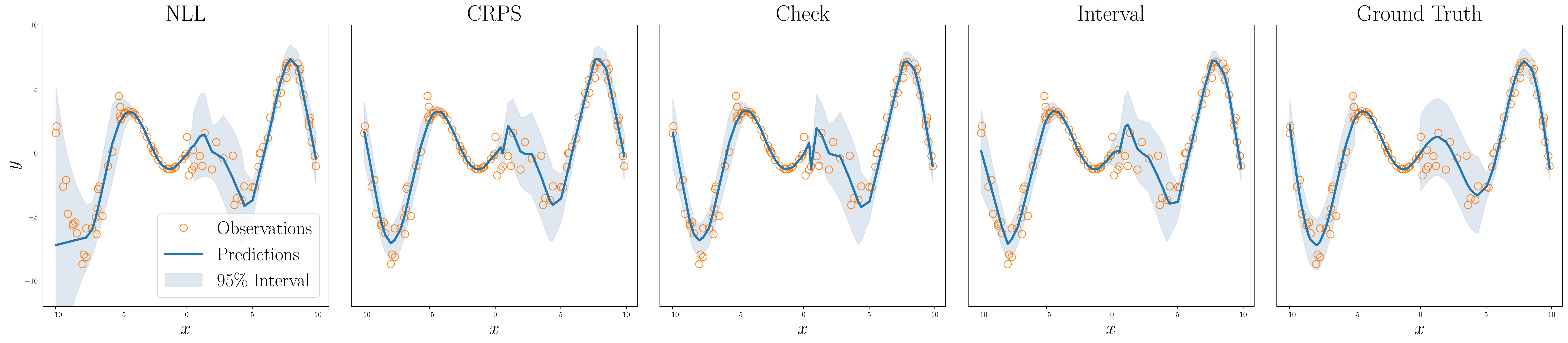}
    \includegraphics[width=\textwidth]{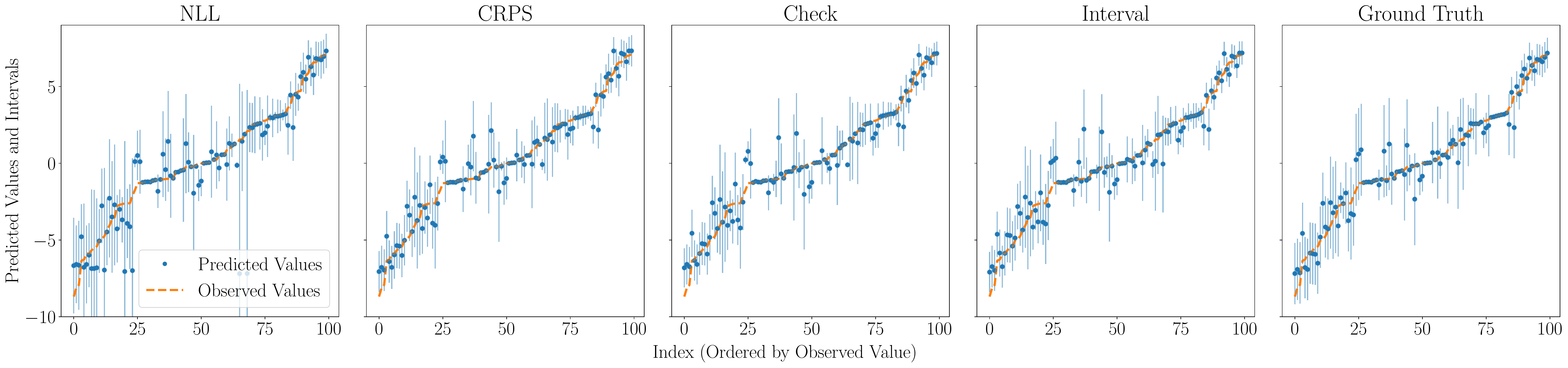}
    \includegraphics[width=\textwidth]{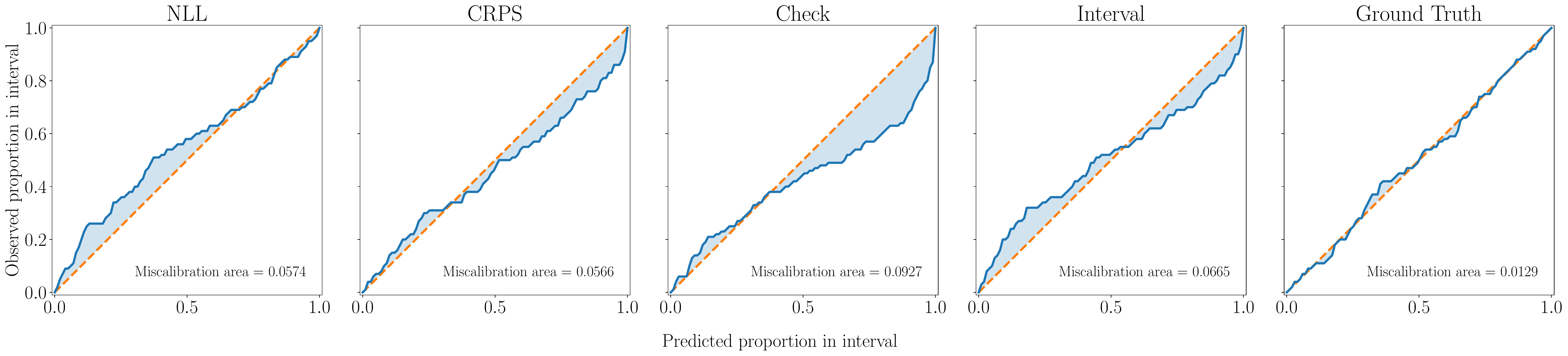}
    \includegraphics[width=\textwidth]{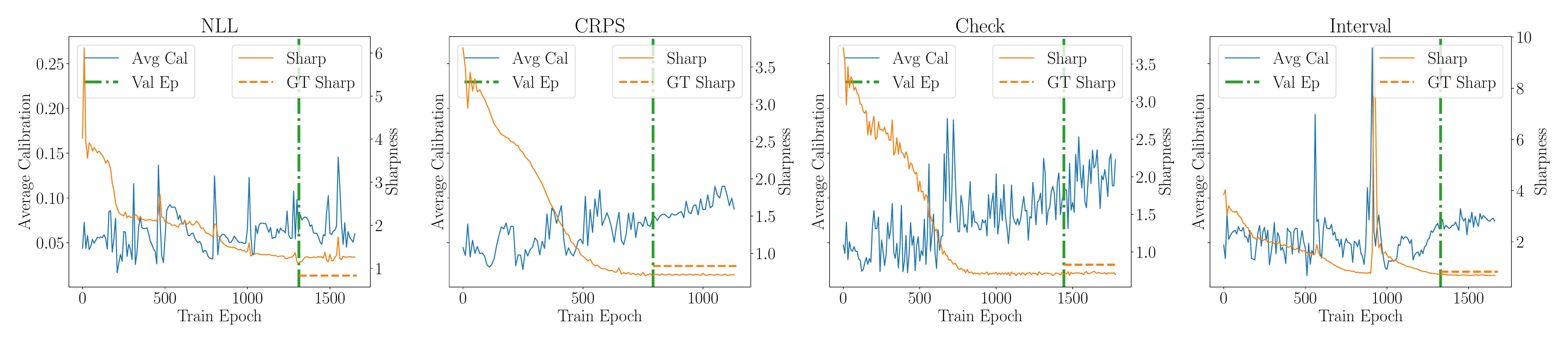}
    \vspace{-8mm}
    \caption{Rows from top to bottom: \textbf{(1)} Test observations, with predicted mean and confidence bands. \textbf{(2)} Test observations, the predicted mean, and prediction interval, in order of test observations. \textbf{(3)} Average calibration plot, with predicted proportions (\textit{expected probability}) on $x$ axis, observed proportions (\textit{observed probability}) on $y$ axis.
    \textbf{(4)} Training curves: {\color{Cerulean} average calibration}  (left $y$ axis), {\color{BurntOrange} sharpness} (right $y$ axis).
    {\color{BurntOrange} GT Sharp} denotes the true sharpness (noise level) of the data,
    and {\color{ForestGreen} Val Ep} denotes the epoch with lowest validation loss.
    }
    \label{fig:plots}
\end{figure*}

\vspace{-4mm}
\paragraph{Analysis} We first visually observe UQ performance on the test set.
Figure~\ref{fig:plots} (row 1) shows all of the methods
approximately recovering the true level of heteroscedastic noise. 
Notably, NLL converges to a solution s.t. for $x < -5$, there is high error in mean estimation, which is compensated for
with high (and wrong) variance estimates.
The widths of the prediction intervals (PIs) in Figure~\ref{fig:plots} (row 2) also show how NLL is erroneously too wide. Meanwhile, comparisons with the ground truth PIs (far right plot) show that CRPS, Check, and Interval 
all tend to have too narrow (sharp) predictions.
This is further confirmed via the \textit{Sharpness} metric column in Table~\ref{table:results}, and we can also 
observe the ramifications in \textit{average calibration} 
in Figure~\ref{fig:plots} (row 3):
NLL's observed proportions in an interval tend to be greater than the
expected proportion, signaling
under-confidence (i.e. PIs that are too wide).
The opposite case occurs for the other methods, and over-confidence (due to PIs that are too sharp) is especially pronounced in CRPS and Check.
While NLL may seem average calibrated (with second lowest ECE),
adversarial group calibration in Figure~\ref{fig:agc}
shows that CRPS and Interval are better calibrated 
for smaller subsets of the domain, and achieve better
adversarial group calibration.

While the proper score metrics in Table~\ref{table:results}
add another facet to the analysis, 
they also underscore the complex nature of assessing UQ.
Each proper score has its own, separate ranking of the four methods, 
and they are also split on which one is best; 
simply given this set of proper scoring rules, we believe 
it would be difficult to choose a single best method.
Lastly, we note how a lower proper score may not necessarily
indicate better calibration (Figure~\ref{fig:plots} (row 4)).
Even while the proper scores improve on the test set (until around the validated epoch), calibration tends to get worse,
while the predictions get sharper. 
Notably, CRPS and Check converge to a solution 
which is sharper than the true sharpness. This is problematic for calibration since a UQ sharper than the true sharpness will never be calibrated.

\vspace{-2mm}
\paragraph{Conclusion}
This case study demonstrates that, even with 
numerous evaluation metrics at our disposal, 
the analysis of UQ for 
regression problems may not be straightforward.
It also highlights limitations of the evaluation metrics, 
as relying on a single one (or small subset),
may imply a conclusion counter to what other metrics signal.
In the face of such limitations, we believe 
it is important to examine a suite of metrics simultaneously and
perform a holistic evaluation of UQ quality.
Not only does Uncertainty Toolbox provide this functionality,
but it also offers recalibration for pre-trained UQ models, and resources
that give key terms, explanations, and seminal works for those unfamiliar with the field.
We hope that this toolbox is useful for accelerating and 
uniting efforts for uncertainty in machine learning.


\newpage


\begin{table*}[h]
\centering
\vspace{2mm}
\begin{tabular}{*{10}{c}} 
\cline{3-10} 
&& \multicolumn{8}{c}{\textbf{Metrics}} \\ 
\cline{3-10} 
& \multicolumn{1}{c}{} & RMSE & MAE & ECE & Sharpness & NLL & CRPS & Check & Interval \\ 
\hline 
\multirow{5}{*}{\textbf{Methods}}
& NLL & $1.689$  & $0.852$  & $0.057$  & $1.451$  & $2.214$  & $0.604$  & $0.305$  & $2.990$  \\
& CRPS       & $\mathbf{0.864}$  & $0.568$  & $\mathbf{0.056}$  & $0.729$  & $1.266$  & $\mathbf{0.427}$  & $\mathbf{0.215}$  & $2.323$  \\
& Check      & $0.880$  & $\mathbf{0.566}$  & $0.092$  & $\mathbf{0.720}$  & $4.264$  & $0.434$  & $0.219$  & $2.434$  \\ 
& Interval   & $0.916$  & $0.600$  & $0.066$  & $0.722$  & $\mathbf{0.780}$  & $0.447$  & $0.226$  & $\mathbf{2.309}$  \\ 
\hhline{~=========}
& Ground Truth   & $\mathit{0.824}$   & $\mathit{0.530}$   & $\mathit{0.013}$   & $\mathit{0.831}$   & $\mathit{-0.083}$   & $\mathit{0.370}$   & $\mathit{0.187}$   & $\mathit{1.758}$  \\ \hline 
\end{tabular}
\caption{\textbf{Scalar Evaluation Metrics.} Each row shows evaluations metrics for a single method (i.e. loss function). RMSE (root mean squared error) and MAE (mean absolute error) are accuracy metrics.
The best method for each metric is in \textbf{bold}.
While these values are based on one seed, 
we show results across 5 random seeds with standard error in Appendix~\ref{app:app_table}.}
\label{table:results}
\end{table*}

\clearpage
\newpage
\bibliographystyle{plainnat}
\bibliography{refs}

\newpage
\appendix
\onecolumn
\section*{Appendix}
\label{sec:appendix}

\section{Details of the Case Study Experiment}
\label{app:exp_details}

\subsection{Details on Dataset}
The synthetic dataset in Section~\ref{sec:case_study} was created with a mean function $y = \sin(x/2) + x\cos(0.8x)$ with $x\sim \text{unif}[-10, 10]$.
The support $[-10, 10]$ was partitioned into 4 quadrants, and different levels of 0 mean, Gaussian noise was added to the mean function to create the $y$ observations.
\begin{align*}
    -10\leq  x  <  -5&    \text{:  noise} \sim\mathcal{N}(0, 1^2) \\
     -5\leq  x  <   0&    \text{:  noise} \sim\mathcal{N}(0, 0.01^2) \\
      0\leq  x  <   5&    \text{:  noise} \sim\mathcal{N}(0, 1.5^2) \\
      5\leq  x \leq 10&   \text{:  noise} \sim\mathcal{N}(0, 0.5^2)
\end{align*}


\subsection{Model Details}
We used the same neural network architecture across all methods (i.e. loss functions): 
3 layers of 64 hidden units with ReLU non-linearities,
and 2 output units: one for the conditional mean $\hat{\mu}(x)$
and one for the conditional log-variance $\log\hat{\sigma^{2}}(x)$.
We used the same learning rate $1e^{-3}$ and full batch size ($200$) for all methods. 
During training, we track the corresponding loss function on the validation set, and at the end of 2000 epochs,
the final model was backtracked to the model with lowest validation loss.
All reported test results are based on this backtracked model.

\subsection{Calculation of Evaluation Metrics}
This section describes how each of the reported metrics is
computed within Uncertainty Toolbox, given a finite dataset $D = \{(x_i, y_i)_{i=1}^{N}\}$.

\textit{Accuracy Metrics}\\
The root mean squared error (RMSE) and mean absolute error (MAE)
are computed with the mean prediction $\hat{\mu}(x)$, following the standard definitions.
\[
\text{RMSE}(D, \hat{\mu}) = \sqrt{\frac{1}{N}\sum_{i=1}^{N}(y_i- \hat{\mu}(x_i))^{2}}
\]
\[
\text{MAE}(D, \hat{\mu}) = \frac{1}{N}\sum_{i=1}^{N}\left|y_i- \hat{\mu}(x_i)\right|
\]

\textit{Calibration Metrics}\\
To measure the calibration metrics (average calibration, adversarial group calibration),
expected probabilities are discretized from 
$0.01$ to $0.99$ in $0.01$ increments 
(i.e. $0.01$, $0.02$, $\dots$, $0.97$, $0.98$, $0.99$).
and the observed probabilities are calculated for each of these $99$ expected probabilities.

ECE (measure of average calibration) is computed following the definition given in Section~\ref{sec:metrics}.

The procedure in which we measure adversarial group calibration is the following.
For a given test set, we scale group size between $1\%$ and $100\%$ of the full test set size, in 10 equi-spaced intervals.
With each group size, we draw 20 random groups from the test set and record the worst calibration incurred across these 20 random groups. 
The adversarial group calibration figure (Figure~\ref{fig:agc}) 
plots the mean worst calibration incurred with $\pm1$ standard error in shades, for each group size. 
This is also the method used by \citet{zhao2020individual} to measure adversarial group calibration.

\textit{Sharpness}\\
Sharpness is measured as the mean of the standard deviation predictions on the test set.
Note that sharpness is a property of the prediction \textit{only} and does not take into consideration the true distribution.

\textit{Proper Scoring Rules}\\
The proper scoring rules (NLL, CRPS, check score, interval score) are measured as the mean of the score on the test set.

\section{Numerical Results Across Multiple Trials}
\label{app:app_table}

The results presented in Section~\ref{sec:case_study} are based on one random seed.
Below, we present the numerical results across 5 random seeds: $[0,1,2,3,4]$.

\begin{table*}[h]
\vspace{2mm}
\centering
\begin{center}
\begin{tabular}{*{6}{c}} 
\cline{3-6} 
&& \multicolumn{4}{c}{\textbf{Metrics}} \\ 
\cline{3-6} 
& \multicolumn{1}{c}{} & RMSE & MAE & ECE & Sharpness \\ 
\hline 
\multirow{5}{*}{\textbf{Methods}}
& NLL       & $2.048 \pm 0.125$   & $1.073 \pm 0.080$   & $\mathbf{0.029} \pm 0.007$   & $1.746 \pm 0.155$   \\
& CRPS      & $\mathbf{1.023} \pm 0.090$   & $\mathbf{0.661} \pm 0.054$   & $0.044 \pm 0.005$   & $0.897 \pm 0.114$   \\
& Check     & $1.045 \pm 0.105$   & $0.672 \pm 0.065$   & $0.050 \pm 0.011$   & $\mathbf{0.874} \pm 0.117$   \\ 
& Interval  & $1.169 \pm 0.187$   & $0.745 \pm 0.101$   & $0.039 \pm 0.009$   & $0.915 \pm 0.130$   \\ 
\hhline{~=====}
& Ground Truth   & $\mathit{0.962 \pm 0.064}$   & $\mathit{0.618 \pm 0.042}$   & $\mathit{0.019 \pm 0.002}$   & $\mathit{0.925 \pm 0.052}$   \\ \hline 
\end{tabular}
\end{center}
\vspace{5mm}
\centering
\vspace{2mm}
\begin{tabular}{*{6}{c}} 
\cline{3-6} 
&& \multicolumn{4}{c}{\textbf{Metrics}} \\ 
\cline{3-6} 
& \multicolumn{1}{c}{} & NLL & CRPS & Check & Interval \\ 
\hline 
\multirow{5}{*}{\textbf{Methods}}
& NLL       & $1.677 \pm 0.343$   & $0.766 \pm 0.060$   & $0.386 \pm 0.030$   & $3.885 \pm 0.330$  \\
& CRPS      & $1.112 \pm 0.111$   & $\mathbf{0.492} \pm 0.040$   & $\mathbf{0.248} \pm 0.020$   & $\mathbf{2.687} \pm 0.186$  \\
& Check     & $1.635 \pm 0.661$   & $0.501 \pm 0.048$   & $0.253 \pm 0.024$   & $2.741 \pm 0.224$  \\ 
& Interval  & $\mathbf{0.961} \pm 0.062$   & $0.546 \pm 0.073$   & $0.276 \pm 0.037$   & $2.875 \pm 0.352$  \\ 
\hhline{~=====}
& Ground Truth   & $\mathit{0.187 \pm 0.115}$   & $\mathit{0.435 \pm 0.033}$   & $\mathit{0.219 \pm 0.017}$   & $\mathit{2.122 \pm 0.177}$  \\ \hline 
\end{tabular}
\caption{\textbf{Scalar Evaluation Metrics.} Each row shows evaluation metrics for a single method (i.e. loss function), and the mean with $\pm1$ standard error are shown. The best mean for each metric has been \textbf{bolded}.}
\label{table:}
\end{table*}

\clearpage
\clearpage

\end{document}